# REAL TIME SPECKLE IMAGE DE-NOISING

D.Sachin Kumar[1], P.R.Seshadri[1], N.Vaishnav[1] and Dr.Saraswathi Janaki[*]

[1]Department of Electronics and Communication Engineering, Easwari Engineering College, Chennai, India
*Professor, Department of Electronics and Communication Engineering, Easwari Engineering College, Chennai, India
`sachin.sky@gmail.com, seshadri8796@gmail.com, vaishnav.chin@gmail.com, saraswathijanakis@gmail.com`

## ABSTRACT

*The paper presents real time speckle de-noising based on activity computation algorithm and wavelet transform. Speckles arise in an image when laser light is reflected from an illuminated surface. The process involves detection of speckles in an image by obtaining a number of frames of the same object under different illumination or angle and comparing the frames for the granular computation and de-noising the same on presence of greater activity index. The project can be implemented in FPGA (Field Programmable Gate Array) technology. The results can be shown that the used activity computation algorithm and wavelet transform has better accuracy in the process of speckle detection and de-noising.*

## 1. INTRODUCTION

Speckles that are present in image processing applications prove to be a challenge in improving the accuracy of image processing. The speckles degrade the quality of image and when it deals with medical oriented applications, it could prove fatal as minute miscalculations in medical imaging may lead to complexions that could prove to be fatal. There is an at most need for detection of speckles in real time and de-noising the same for improved processing. The Granular computation and activity computation algorithm are used to bring about speckle detection.

This process involves capturing of multiple frames of the same image with difference in the illumination or different angle. Each frame contains same resolution in means of the number of pixels. These frames are compared successively to obtain the granular count and further used to determine the activity index. This activity index is the measure of the speckle that is related to the image. Based upon a threshold value of the activity index the image is forwarded for de-noising.

## 2. ACTIVITY INDEX COMPUTATION

The data source consists of $N_F$ number of frames, each frames consisting of Np number of pixels in total.

$N_p = N_C \times N_r$

Where $N_C$ and $N_r$ are number pixels in each column and row of the frame.

The following algorithm consists of three steps. First two steps are actually preset procedures, the first step sets the intensity histogram, and the second step computes the actual region limits of the distribution of intensity levels. The third step computes the granular count and the activity index.

## 2.1. Obtaining Histogram

The histogram is a plot between the gray level values and probability of occurrence of pixel. It is the basic step that needs to be performed to get an overview of the gray level intensity value distribution. Based upon this distribution, the next step is performed. For counting the number of pixels bin registers are used for each gray value.

## 2.2. Finding Region Boundaries

Depending upon the number of regions that are required for effective analysis of the frame, the region limits are determined based upon the number of pixels that fall under that range of gray level intensity values in each region. It is essential that each region should comprise of almost equal number of pixels.

Let the total range be from [0,255], to determine a region two values are required, one is the upper limit and other is the lower limit. If there are Z regions given then 2Z-1 regions are produced due to overlapping of the successive regions. This ensures that entire ranges of gray level values are covered and provides more clarity for upcoming procedures. the number of pixels that each region must contain approximately is given as

$$N_m = N_p / (2Z-1)$$

Let bin (i) be the last on to be included in the first counting, (i) will be the upper limit of the first region, and
(i+1) the lower limit of the next region. This is then repeated from bin (i+1) up to the point when the accumulated count is closest to Nm. This way, all the region limits will be computed.

## 2.3. Estimation of Granular Count:

At the granular computing stage all the frames are lined up sequentially in such a way that the corresponding pixel positions of the successive frames can be accessed serially.
For each incoming pixel the intensity values is verified and determined under which region it falls and it is compared with same pixel in the previous frame. The granular count value is initially set to be zero. If the current pixel's region is different from that of the corresponding pixel in the previous frame then the granular count value is incremented by one. If the regions of the successive pixels happen to be in same region then the granular count value is unperturbed and kept the same until the next region change occurs. The successive pixels of the successive frames that falling under the same region limit is said to be on granule. The number of granules that is formed in every region is counted, that is represented as the granular count.
From this granular count activity index is calculated by dividing the total number granules by number of frames used for computation.

## 2.4. Memory Requirements

Granule counters: $N_p = N_r \times N_c$ granule counters are needed to handle the pixel activity which requires a memory of

$$C_{mem} = N_p \times L$$

Where L is the register-cell size.

Histogram setup: considering the intensity values in the range [0,255]; 256 bin-memory registers are needed to accommodate all the pixels.

$$H_{mem} = 256 \times \log N_p / \log 2$$

.

## 3. SETTING THRESHOLD

Based upon the application and the accuracy that is desired a threshold value of the activity index is set.

After the computation of the activity index is done, it is compared with the already set threshold value and it decides whether the image is affected by unacceptable level of speckle and whether it requires to be denoised. If the image needs to be de-noised it is sent to the denoising part for further processing, else the process ends stating that the image is free from speckle.

## 4. DE-NOISING BY WAVELET TRANSFORM

A wavelet series is a representation of a square-integral (real- or complex-valued) function by a certain orthonormal series generated by a wavelet. Nowadays, wavelet transformation is one of the most popular candidates of the time-frequency-transformations. This article provides a formal, mathematical definition of an orthonormal wavelet and of the integral wavelet transform.

Using this wavelet transform the image that contains speckle can be de-noised. The type of wavelet used here is 'haar' wavelet.

## 5. HARDWARE IMPLEMENTATION

### 5.1. Histogram module

The histogram is first calculated by the histogram hardware module. Here each pixel value is given as input sequentially to the module and it stores them in a memory register. The plot of the pixel intensity and the probability of occurrence give the histogram of the image. The module is implemented using HDL (Hardware Description Language).

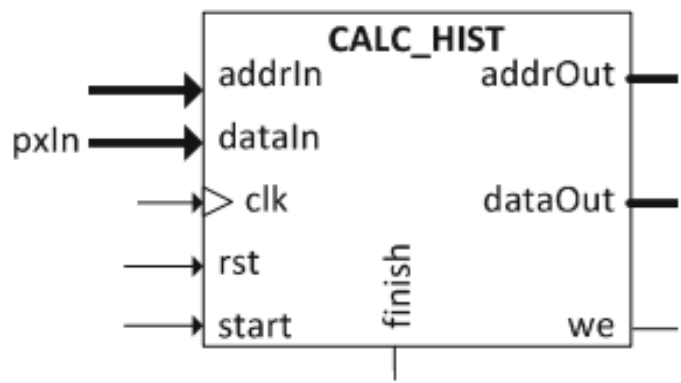

On each clock cycle the pixels (pxIn) are given as input in sequence. The histogram is stored in MEM_HIST register. Finish is used to denote that the histogram calculation module is finished.

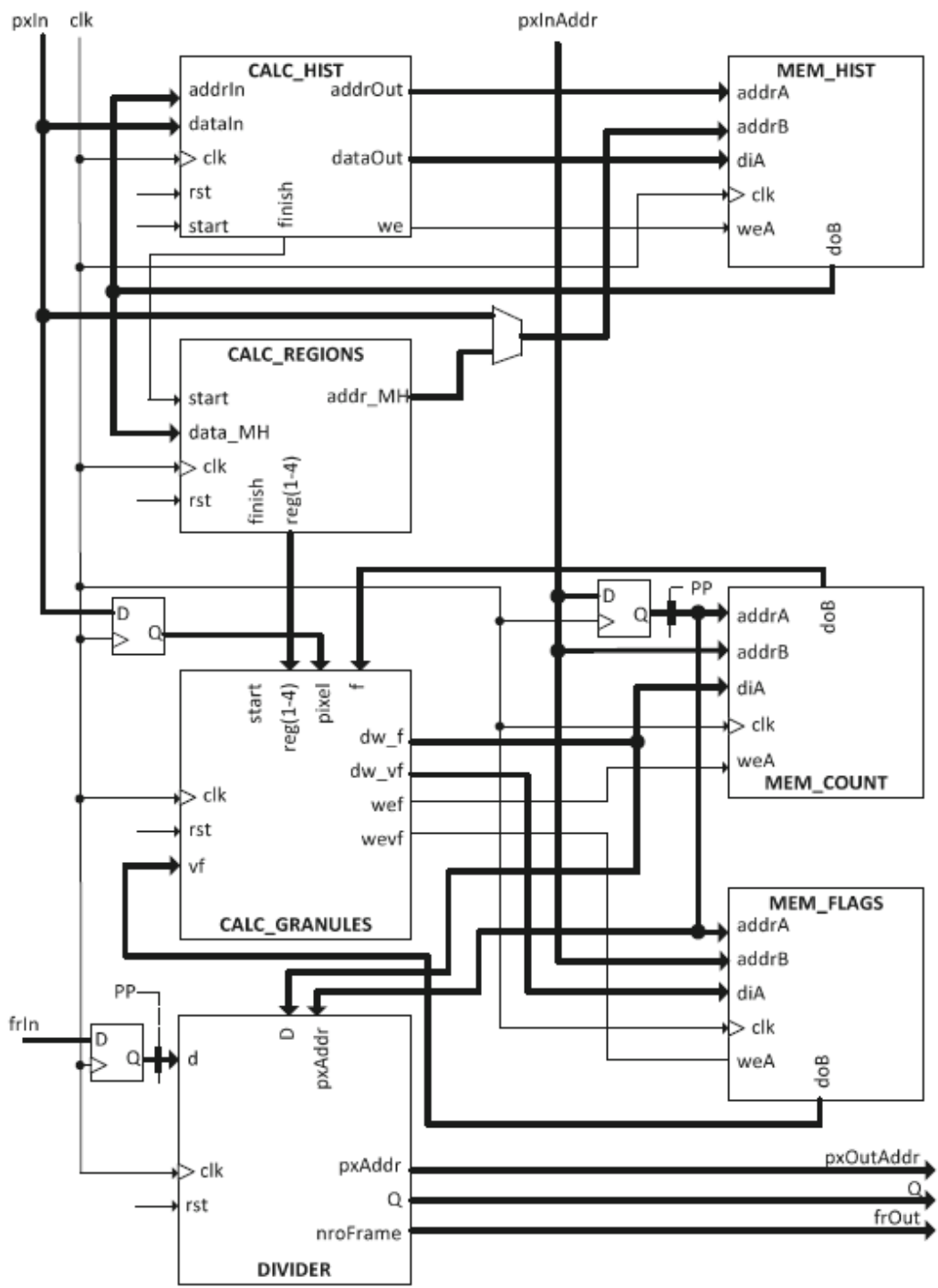

Fig1 : Digital Circuit to Implement activity index

## 5.2. Region Computation module

The region limits are computed by the region computation module. The required number of regions is set and the regions are split such that equal numbers of pixels are present in each region. The module is implemented using HDL coding

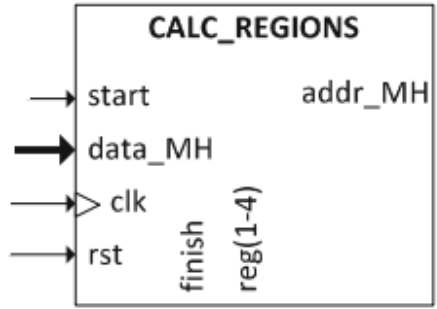

The variable reg value can range from 1 to 4. The boundary value of the regions gets stored in the memory for processing

## 5.3. Granular Computation module

The granular count is calculated using the granular computation module generated using the HDL coding. It calculated the total number of pixels that are different in position in successive frames and the final obtained value for a 512x512 image is computed.

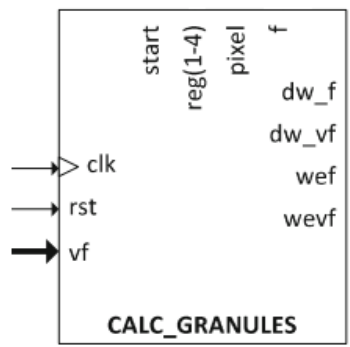

The granular count value gets stored in MEM_FLAGS for next step of processing.

## 5.4. Activity index module

The activity index is calculated using the activity index module generated using the HDL coding. It consists of divider unit that divides the granular count to the number of frames.
Activity index= (granular count/number of frames)

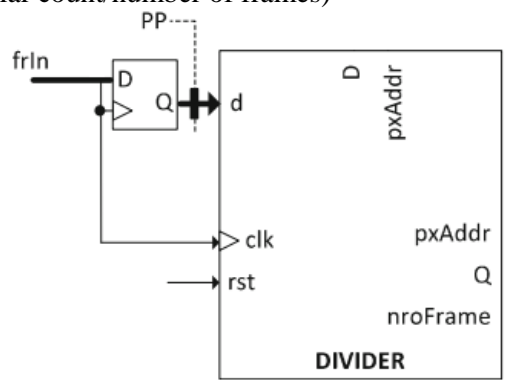

The activity index threshold value needs to be fixed and is found out by trial and error method.

## 5.5. De-noising module

The de-noising is done if the activity index is greater than the threshold value. The de-noising module is generated by converting the Matlab coding into HDL using the code generation module of Matlab.

If the Activity index value is less than the threshold value, then there is no need to de-noise the image as the image has negligible noise and the next set of frames is obtained for processing from the first stage.

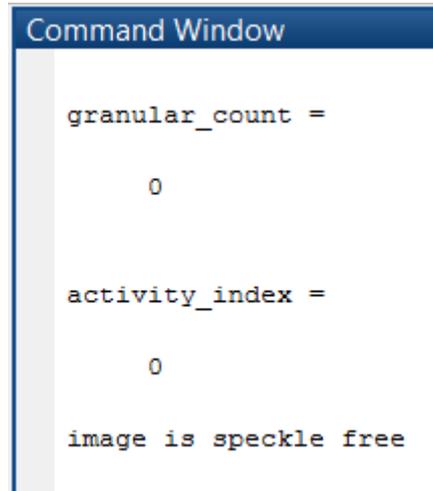

The above result shows that the image is speckle free as there is no noise in the image.

## 5.5. Tabulation of Hardware results

| VARIANCE | MSE1 | MSE2 | PSNR1 | PSNR2 | IEF |
|---|---|---|---|---|---|
| 0.08 | 0.047 | 0.048 | 46.23 | 41.73 | 2.83 |
| 0.001 | 0.017 | 0.027 | 49.75 | 50.78 | 1.78 |

**VARIANCE** – Amount of noise variance added to original image during initial processing stage

**MSE1** – Mean square error between original noiseless image and noisy image obtained.

**MSE2** - Mean square error between noisy image and the image obtained after de-noising by means of wavelet transform.

**PSNR1** – Peak Signal to Noise Ratio between original noiseless image and noisy image obtained.

**PSNR2** - Peak Signal to Noise Ratio between noisy image and the image obtained after de-noising by means of wavelet transform.

**IEF** – Image Enhancement Factor

## 6. RESULTS & DISCUSSION:

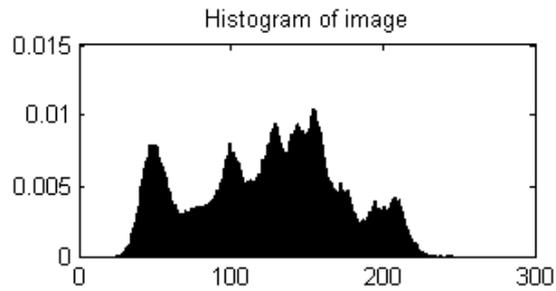

(i)

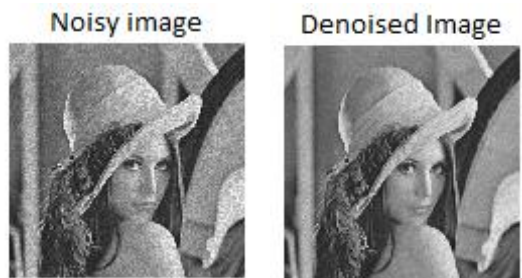

(ii)                    (iii)

(i) Histogram of the image
(ii) Noisy image obtained by adding speckle noise to the original noise free image
(iii) Image de-noised using wavelet transform

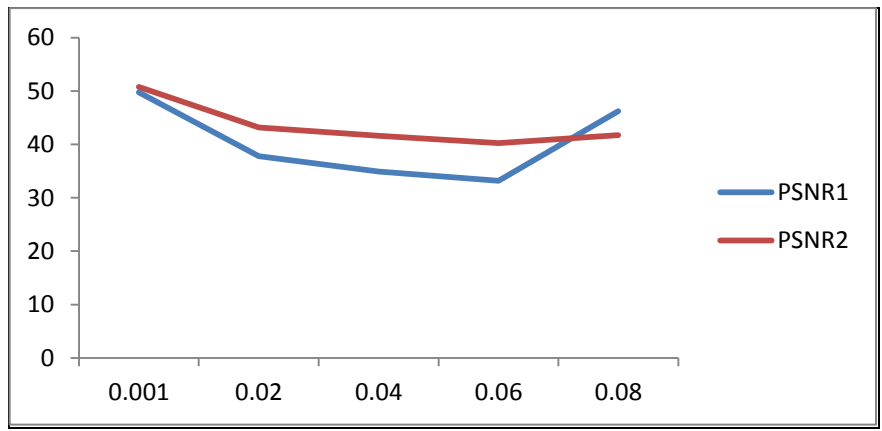

Fig 2: Noise Variance vs PSNR

From the above graph it is seen that the PSNR value tends to be high and almost constant for various values of noise variance. The PSNR values seems to be least affected by the increase in the noise variance. This proves to be a consistent and reliable algorithm for speckle denoising.

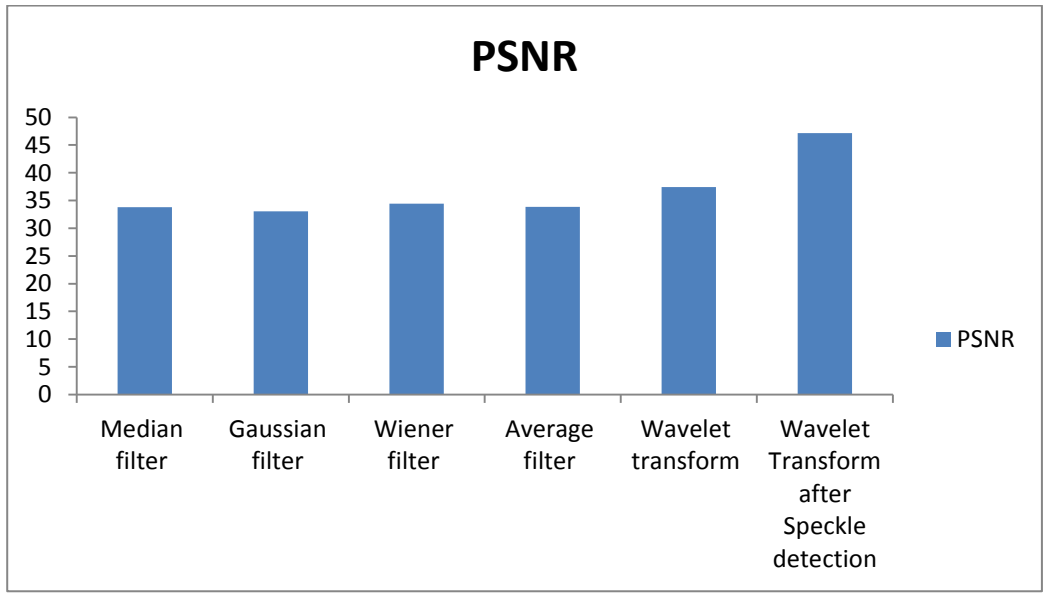

**Comparison for PSNR values of various filters**

From the above graph, it is evident that the PSNR is higher for the designed algorithm which involves detection of speckle and de-noising the speckle on its presence. The proposed algorithm thereby saves time and is also much efficient compared to other algorithms.\

## 7. CONCLUSIONS

The use of activity index computation enables us to detect speckle and de-noise efficiently and accurately. This is a unique algorithm to detect the speckle which will be easier when it comes to real time implementation.

This algorithm proves to be effective in analysis and detecting even a minimal amount of speckle that is involved in the images. It is dynamic in nature which is capable of operating with images of better resolution apart from the low resolution images.

### ACKNOWLEDGEMENTS

I would like to thank everyone who supported for this Paper. Especially I would like to thank my Guide Dr. Saraswathi Janaki who helped for this project and insisted to extend the Project by extending valuable ideas.